\documentclass[10pt,twocolumn]{article}

\usepackage{cvpr}

\usepackage{times}
\usepackage{epsfig}
\usepackage{graphicx}
\usepackage{amsmath}
\usepackage{amssymb}
\usepackage{rotating}
\usepackage{color}
\usepackage{framed}
\usepackage{balance}

\def \ie {\emph{i.e.}}
\def \eg {\emph{e.g.}}
\def \etal {\emph{et al.}}

\definecolor{darkgreen}{RGB}{0, 140, 0}
\newcommand{\rv}[1]{{\color{darkgreen}[\textbf{Rv}:#1]}}
\newcommand{\arthur}[1]{{\color{red}[\textbf{Arthur}:#1]}}
\newcommand{\matthijs}[1]{{\color{blue}[\textbf{Matthijs}:#1]}}

\usepackage[pagebackref=false,breaklinks=true,letterpaper=true,colorlinks,bookmarks=false]{hyperref}

\def\cvprPaperID{1590} %

\cvprfinalcopy

\ifcvprfinal
	\def\httilde{\mbox{\tt\raisebox{?}{?}x{-.5ex}{\symbol{126}}}}
\fi

\def \W {\mathbf{W}} 
\def \L {\mathbf{L}} 

\newcommand{\ndis}{n_\mathrm{B}}
\newcommand{\nlab}{n_\mathrm{L}}
\newcommand{\std}[1]{\footnotesize{$\pm$#1}}

\begin{document}

\title{Low-shot learning with large-scale diffusion}

\author{Matthijs Douze\textsuperscript{$\dagger$}, 
Arthur Szlam\textsuperscript{$\dagger$}, 
Bharath Hariharan\textsuperscript{$\dagger$}\thanks{This work was carried out while B. Hariharan was post-doc at FAIR.}~, 
Herv\'e J\'egou\textsuperscript{$\dagger$}\\
\textsuperscript{$\dagger$}Facebook AI Research\\
\textsuperscript{*}Cornell University\\
}

\def \OLD  {\textcolor{red}{\bf OLD}}

\maketitle
\begin{abstract}
This paper considers the problem of inferring image labels from images when only a few annotated examples are available at training time. This setup is often referred to as low-shot learning, where a standard approach is to re-train the  last few layers of a convolutional neural network learned on separate classes for which training examples are abundant. 
We consider a semi-supervised setting based on a large collection of images to support label propagation.  This is  possible by leveraging the recent advances on large-scale similarity graph construction. 

We show that despite its conceptual simplicity, scaling label propagation up to hundred millions of images leads to state of the art accuracy in the low-shot learning regime. 
\end{abstract}

\section{Introduction}

Large, diverse collections of images are now commonplace; these often contain a ``long tail'' of visual concepts.  Some concepts like ``person'' or ``cat'' appear in many images, but the vast majority of the visual classes do not occur frequently.   Even though the total number of images may be large, it is hard to collect enough labeled data for most of the visual concepts.  Thus if we want to learn them, we must do so with few labeled examples. This task is named \emph{low-shot learning} in the  literature. 

In order to learn new classes with little supervision, a standard approach is to leverage classifiers already learned for the most frequent classes, employing a so-called \emph{transfer learning} strategy. For instance, for new classes with few labels, only the few last layers of a convolutional neural network are re-trained. This limits the number of parameters that need to be learned and limits over-fitting. 

In this paper, we consider the low-shot learning problem described above, where the goal is to learn to detect new visual classes with only a few annotated images per class, but we also assume that we have many unlabelled images. This is called semi-supervised learning~\cite{ZG2009,zhou2003learning} (considered, \eg, for face annotation~\cite{FWT09}).  The motivation of this work is threefold.   First we want to show that with modern computational tools, classical semi-supervised learning methods scale gracefully to hundreds of millions of unlabeled points.  A limiting factor in previous evaluations was that constructing the similarity graph supporting the diffusion was slow. This is no longer a bottleneck: thanks to advances both in computing architectures and algorithms, one can routinely compute the similarity graph for 100 millions images in a few hours~\cite{johnson2017billion}.  Second, we want to answer the question: \emph{Does a very large number of images help for semi-supervised learning?} %
Finally, by comparing the results of these methods on Imagenet and the YFCC100M dataset~\cite{TFENPBL16}, we highlight how these methods exhibit some artificial aspects of Imagenet that can influence the performance of low shot learning algorithms.

\begin{figure}[t]
\begin{minipage}{\linewidth}
\includegraphics[width=1.0\linewidth]{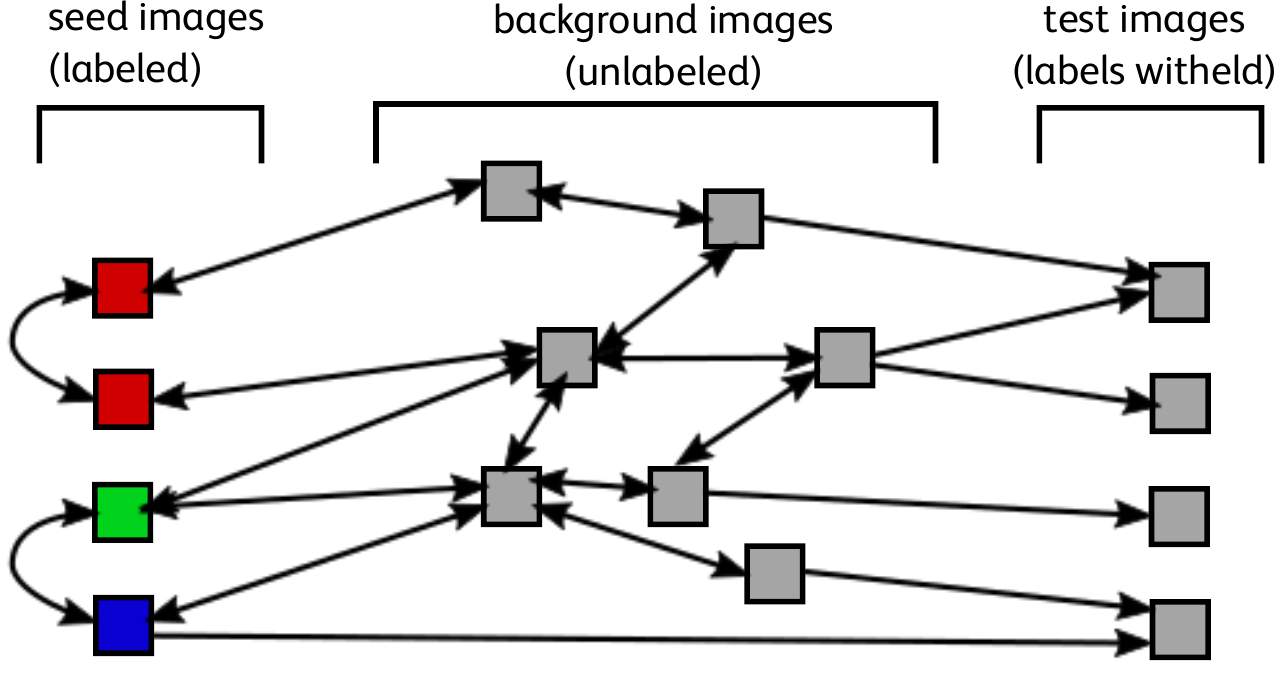}
\end{minipage}
\caption{\label{fig:graphdiffusion}
	The diffusion setup. The arrows indicate the direction of diffusion. There is no diffusion performed from the test images. For the rest of the graph, the edges are bidirectional (\ie, the graph matrix is symmetric). Except when mentioned otherwise, the edges have no weights.
}
\end{figure}

In summary, the contribution of our paper is a study of semi-supervised learning  in the scenario where we have a very large number of unlabeled images.
Our main results are that in this setting, semi-supervised learning leads to state of the art low-shot learning performance.
 In more detail, we make the following contributions:
\begin{itemize}
\item 
We carry out a large-scale evaluation for diffusion methods for semi-supervised learning and compare it to recent low-shot learning papers. Our experiments are all carried out on the public benchmark Imagenet~\cite{DSLLF09} and the YFC100M dataset~\cite{TFENPBL16}.  
\item We show that our approach is efficient and that the diffusion process scales up to hundreds of millions of images, which is order(s) of magnitude  larger than what we are aware in the literature on image-based diffusion~\cite{iscen2017efficient,iscen2017fast}. 
This is made possible by leveraging the recent state of the art for efficient k-nearest neighbor graph construction~\cite{johnson2017billion}.  
\item We evaluate several variants and hypotheses involved in diffusion methods, such as using class frequency priors~\cite{ZGL03}. This scenario is realistic in  situations where this statistic is known a priori. We propose a simple way to estimate it without this prior knowledge, and extend this assumption to a multiclass setting by introducing a probabilistic projection step derived from Sinkhorn-Knopp algorithm. 
\item Our experimental study shows that a simple propagation process significantly outperforms some state-of-the-art approaches in low-shot visual learning when (i) the number of annotated images per class is small and when (ii) the number of unlabeled images is large or the unlabeled images come form the same domain as the test images. 
\end{itemize}

This paper is organized as follows. Section~\ref{sec:related} reviews related works and Section~\ref{sec:propagation} describes the label propagation methods. The experimental study is presented in Section~\ref{sec:experiments}. Our conclusion in section~\ref{sec:conclusion} summarizes our findings.

\section{Related work}
\label{sec:related}

\paragraph{Low-shot learning}
Recently there has been a renewed interest for low-shot learning, \ie, learning with few examples thanks to prior statistics on other classes. Such works include metric learning~\cite{MVPC12}, learning kNN~\cite{VBLW16}, regularization and feature hallucination~\cite{bharath2017low} or predicting parameters of the network~\cite{BHVTV16}. Ravi and Larochelle introduce a meta-learner to learn the optimization parameters invovled in the low-shot learning regime~\cite{RL17}. 
Most of the works consider small datasets like Omniglot, CIFAR, or a small subset of Imagenet. In our paper we will focus solely on large datasets, in particular the Imagenet collection~\cite{ILSVRC15} associated with the ILSVRC challenge.

\paragraph{Diffusion methods}

We refer the reader to \cite{chapterLabelProp06,DB13} for a review of diffusion processes and matrix normalization options.
Such methods are an efficient way of clustering images given a matrix of input similarity, or a kNN graph, and have been successfully used in a semi-supervised discovery setup~\cite{FWT09}. They share some connections with spectral clustering~\cite{NLCG08}. 
In~\cite{PZ08}, a kNN graph is clustered with spectral clustering, which amounts to computing the $k$ eigenvectors associated with the $k$ largest eigenvalues of the graph, and clustering these eigenvectors. Since the eigenvalues are obtained via Lanczos iterations~\cite[Chapter~10]{GL13}, the basic operation is similar to a diffusion process.
This is also related to power iteration clustering~\cite{LC10}, as in the work of Cho \etal~\cite{CL12} to find clusters. %

\paragraph{Semi-supervised learning}
The kNN graph can be used for transductive and semi-supervised learning (see \eg \cite{chapterLabelProp06,ZG2009} for an introduction).  In transductive learning, a relatively small number of labels are used to augment a large set of unlabeled data and the goal is to extend the labeling to the unlabeled data (which is given at train time).  Semi-supervised learning is similar, except there may be a separate set of test points that are not seen at train time.  
In our work, we consider the simple proposal of Zhu~\etal~\cite{ZGL03}, where powers of the (normalized) kNN graph are used to find smooth functions on the kNN graph with desired values at the labeled points. 
There exist many variations on the algorithms, \eg,  Zhou \etal~\cite{zhou2003learning} weight the edges based on distances and introduce a loss trading a classification fitting constraint and a smoothness term enforcing consistency of neighboring nodes. 

Label propagation is a transductive method. In order to evaluate on new data, we need to 
extend the smooth functions out of the training data.  While deep networks have been used before for out of sample extension, \eg,  in \cite{CWSS07} and \cite{JSL15}, in the speech domain, in this work, we use  a weighted sum of nearest neighbors from the (perhaps unlabeled) training data \cite{BPVDR03}.   %

\paragraph{Efficient kNN-graph construction} The diffusion methods use a matrix as input containing the similarity between all  images of the dataset. Considering $N$ images, \eg, $N=10^8$, it is not possible to store a matrix of size $N^2$. However most of the image pairs are not related and have a similarity close to 0. 
Therefore diffusion methods are usually implemented with sparse matrices. This means that we compute a graph connecting each image to its neighbors, as determined by the similarity metric between image representations. In particular, we consider the k-nearest neighbor graph (kNN-graph) over a set of vectors. 
Several approximate algorithms~\cite{DCL11,KKNMS16,AKAE15,HD16} have been proposed to efficiently produce the kNN graph used as input of iterative/diffusion methods, since this operation is of quadratic complexity in the number of images. 
In this paper, we employ the Faiss library,%
which was shown capable to construct a graph connecting up to 1 billion vectors~\cite{johnson2017billion}.

\section{Propagating labels}
\label{sec:propagation}

This section describes the initial stage of our proposal, which estimates the class of the unlabelled images with a diffusion process. It includes an image description step, the construction of a kNN graph connecting similar images, and a label diffusion algorithm.

\subsection{Image description} 
\label{sec:description}
A meaningful semantic image representation and an associated metric is required to match instances of classes that have not been seen beforehand. %
While early works on  semi-supervised labelling~\cite{FWT09} were using ad-hoc semantic global descriptors like GIST~\cite{OT01}, 
we extract activation maps from a CNN trained on images from a set of \emph{base} classes that are independent from the \emph{novel} classes on which the evaluation is performed. See the experimental section for more details about the training process for descriptors.

The mean class classifier introduced for low-shot learning~\cite{MVPC12} is another way to perform dimensionality reduction while improving accuracy thanks to a better comparison metric. We do not consider this approach since it can be seen as part of the descriptor learning.

\subsection{Affinity matrix: approximate kNN graph}
\label{sec:propagationmatrix}

As discussed in the related work, most diffusion processes use as input the kNN graph representing the $N\times N$ sparse similarity matrix, denoted by $\W$, which connects the $N$ images of the collection. 
We build this graph using approximate k-nearest neighbor search.
Thanks to recent advances in efficient similarity search~\cite{DCL11,johnson2017billion}, trading some accuracy against efficiency drastically improves the graph construction time. As an example, with the \textsc{Faiss} library~\cite{johnson2017billion}, building the graph associated with 600k images takes 2~minutes on 1~GPU. In our preliminary experiments, the approximation in the knn-graph does not induce any sub-optimality, possibly because the diffusion process compensates the artifacts induced by the approximation.

Different strategies exist to set the weights of the affinity matrix $\W$. We choose to search the $k$ nearest neighbors of each image, and set a 1 for each of the neighbors in the corresponding row of a sparse matrix $\W_0$. Then we symmetrize the matrix by adding it to its transpose. We subsequently $\ell_1$-normalize the rows to produce a sparse stochastic matrix: $\W=D^{-1}(\W_0^\top + \W_0)$, with $D$ the diagonal matrix of row sums. %

The handling for the test points is different: test points do not participate in label propagation because we classify each of them independently of the others. Therefore, there are no outgoing edges on test points; they only get incoming edges from their $k$ nearest neighbors. 

\subsection{Label propagation}
\label{sec:labelpropagation}

We now give details about the diffusion process itself, which is summarized in Figure~\ref{fig:graphdiffusion}.
We build on the straightforward label propagation algorithm of ~\cite{ZGL03}. The set of images on which we perform diffusion is composed of  $\nlab$ labelled \emph{seed} images and $\ndis$ unlabelled \emph{background} images ($N=\nlab + \ndis$). Define the $N \times C$ matrix $\L$, where $C$ is the number of classes for which we want to diffuse the labels, \ie, the new classes not seen in the training set.  
Each row $l_i$ in $\L$  is associated with a given image, and represents the probabilities of each class for that image. A given column corresponds to a given class, and gives its probabilities for each image. The method initializes $l_i$ to a one-hot vector for the seeds. Background images are initialized with 0 probabilities for all classes. 
Diffusing from the known labels, the method iterates as %
$\L_{t+1} = \W \L_t$. 
We can optionally reset the $\L$ rows corresponding to seeds to their 1-hot ground-truth at each iteration. 
When iterating to convergence, all $l_i$ would eventually converge to the eigenvector of $\W$ with largest eigenvalue (when not resetting), or to the harmonic function with respect to $\W$ with boundary conditions given by the seeds (when resetting). Empirically, for low-shot learning, we observe that resetting is detrimental to accuracy. Early stopping performs better in both cases, so we cross-validate the number of diffusion iterations.

\paragraph{Classification decision \& combination with logistic regression}
We predict the class of a test example $i$ as the the column that maximizes the score $l_i$. 
Similar to Zhou \etal~\cite{zhou2003learning}, we have also optimized a loss balancing the fitting constraint with the diffusion smoothing term. However we found that a simple late fusion (weighted mean of log-probabilities, parametrized by a single cross-validated coefficient) of the scores produced by diffusion and logistic regression achieves better results. %

\subsection{Variations} 
\label{sec:variations}

\paragraph{Using priors}

The label propagation can take into account several priors depending on the assumptions of the problem, which are integrated by defining a normalization operator $\eta$ and by modifying the update equation as 
\begin{equation} 
\L_{t+1} = \eta (\W \L_t). 
\label{equ:sspic_update}
\end{equation}

\noindent \emph{Multiclass assumption.} For instance, in the ILSVRC challenge built upon the Imagenet dataset~\cite{ILSVRC15}, there is only one label per class, therefore we can define $\eta$ as a function that $\ell_1$-normalizes each row to provide a distribution over labels (by convention the normalization leaves all-0 vectors unchanged).
\medskip 

\noindent \emph{Class frequency priors.}
 Additionally, we point out that labels are evenly distributed in Imagenet. 
If we translate this setup to our semi-unsupervised setting, it would mean that we may assume that the distribution of the unlabelled images is uniform over labels. This assumption can be taken into account by defining $\eta$ as the function performing a $\ell_1$ normalization of columns of~$\L$. 

While one could argue that this is not realistic in general,  a more realistic scenario is to consider that we know the marginal distribution of the labels, as proposed by Zhu \etal~\cite{ZGL03}, who show that the prior can be simply enforced (\ie, apply column-wise normalization to $\L$ and multiply each column by the prior class probability). 
This arises in situations such as tag prediction, if we can empirically measure the relative probabilities of tags, possibly regularized for lowest values.  
\medskip 

\noindent \emph{Combined Multiclass assumption and class frequency priors. }
We propose a variant way to use both a multiclass setting and prior class probabilities by enforcing the matrix $\L$ to jointly satisfy the following properties:
\begin{align}
\L \mathbf{1}_\mathrm{C} & = \mathbf{1}_\mathrm{N}  & 
\mathbf{1}_\mathrm{N}^\top \L & \propto \mathbf{p}_{\mathcal C}
\label{equ:const2}
\end{align}
where $p_{\mathcal C}$ is the prior distribution over labels.
For this purpose, we adopt a strategy similar to that of Cuturi~\cite{cuturi2013sinkhorn} in his work on optimal transport, in which he shows that the Sinkhorn-Knopp algorithm~\cite{S67} provides an efficient and theoretically grounded way to project a matrix so that it satisfies such marginals. The Sinkhorn-Knopp algorithm iterates by alternately enforcing the marginal conditions, as
\begin{align}
\L & \leftarrow \L \  \mathrm{diag}(\L \mathbf{1}_\mathrm{C})^{-1} \mathrm{diag}(\mathbf{p}_{\mathcal C}) \\
\L & \leftarrow \mathrm{diag}(\mathbf{1}_\mathrm{N}^\top \L )^{-1} \L 
\label{equ:up2}
\end{align}
until convergence. Here we assume that the algorithm only operates on rows and columns whose sum is strictly positive. 
As discussed by Knight~\cite{K08}, the convergence of this algorithm is fast. Therefore we stop after 5 iterations. This projection is performed after each update by Eqn.~\ref{equ:sspic_update}. 
Note that Zhu \etal~\cite{ZGL03} solely considered the second constraint in Eqn.~\ref{equ:const2}, which can be obtained by enforcing the prior, as discussed by  Bengio \etal\cite{chapterLabelProp06}. 
We evaluate both variants in the experimental section~\ref{sec:experiments}.

\paragraph{Non-linear updates.}  
The Markov Clustering (MCL)~\cite{EDO2002} is another diffusion algorithm with nonlinear updates originally proposed for clustering. 
In contrast to the previous algorithm,  MCL iterates directly over the similarity matrix as
\begin{align}
\W_t' \leftarrow \W_t \cdot \W_t & & 
\W_{t+1} \leftarrow \Gamma_r(\W_t'), 
\end{align}
where $\Gamma_r$ is an element-wise raising to power $r$ of the matrix, followed by a column-wise normalization~\cite{EDO2002}. The power $r\in (1, 2]$ is a bandwidth parameter: when $r$ is high, small edges quickly vanish along the iterations. A smaller $r$ preserves the edges longer. 
The clustering is performed by extracting connected components from the final matrix.
In Section~\ref{sec:experiments} we evaluate the role of the non-linear update of MCL by introducing the $\Gamma_r$ non-linearity in the diffusion procedure. More precisely, we modify Equation~\ref{equ:sspic_update} as 
$\L_{t+1} = \Gamma_r \left( \eta (\W \L_t) \right). $

\subsection{Complexity}
\label{sec:complexity}

\newcommand{\iterlogreg}{I_\mathrm{logreg}}
\newcommand{\iterdif}{I_\mathrm{dif}}

For the complexity evaluation, we distinguish two stages. In the \textbf{off-line} stage, \emph{(i)} the CNN is trained on the base classes, \emph{(ii)} descriptors are extracted for the background images, and \emph{(iii)} a knn-graph is computed for the background images. In the \textbf{on-line} stage, we receive training and test images from novel classes,  \emph{(i)} compute features for them,  \emph{(ii)} complement the knn-graph matrix to include the training and test images, and  \emph{(iii)} perform the diffusion iterations. 
Here we assume that the $N\times N$ graph matrix $\W_0$ is decomposed in four blocks 
\begin{equation}
\W_0 = \left[
\begin{array}{cc}
\W_\mathrm{LL} & \W_\mathrm{LB} \\
\W_\mathrm{BL} & \W_\mathrm{BB} \\
\end{array}
\right] \in \{0,1\}^{(\nlab+\ndis)\times (\nlab+\ndis)}
\end{equation}
The largest matrix $\W_\mathrm{BB}\in\{0,1\}^{\ndis\times \ndis}$ is computed off-line. On-line we compute the three other matrices. We combine $\W_\mathrm{BL}$ and $\W_\mathrm{BB}$ by merging similarity search result lists, hence each row of $\W_0$ contains exactly $k$ non-zero values, requiring to store the distances along with $\W_\mathrm{BB}$.

We are mostly interested in the complexity of the on-line phase. Therefore we exclude the descriptor extraction, which is independent of the classification complexity, and the complexity of handling the test images, which is negligible compared to the training operations. We consider the logistic regression as a baseline for the complexity comparison:
\begin{description}
\item 
[Logistic regression] the SGD training entails $\mathcal{O}(\iterlogreg \times B \times C \times d)$ multiply-adds, with $d$  denotes the descriptor dimensionality and $C$ the number of classes. The number of iterations and batch size are $\iterlogreg$ and $B$. 
\item 
[Diffusion] the complexity is decomposed into: computing the matrices $\W_\mathrm{LL}$, $\W_\mathrm{LB}$ and $\W_\mathrm{BL}$, which involves $\mathcal{O}(d \times \nlab \times \ndis)$ multiply-adds using brute-force distance computations; and performing $\iterdif$ iterations of sparse-dense matrix multiplications, which incurs $\mathcal{O}(k \times N \times C \times \iterdif)$ multiply-adds (note, sparse matrix operations are more limited by irregular memory access patterns than arithmetic operations). Therefore the diffusion complexity is linear in the number of background images $\ndis$. 
See the supplemental for more details. 
\end{description}

\paragraph{Memory usage.}

One important bottleneck of the algorithm is its memory usage. The sparse matrix $\W_0$ occupies $8 N k$ bytes in RAM, and $\W$ almost twice this amount, because most nearest neighbors are not reciprocal; the $\L$ matrix is $4 C N$ bytes. Fortunately, the iterations can be performed one column of $\L$ at a time, reducing this to $2\times 4N$ bytes for $\L_t$ and $\L_{t+1}$ (in practice, when memory is an issue, we group columns by batches of size $C' < C$). 

\section{Experiments}
\label{sec:experiments}

\newcommand{\ig}[1]{\includegraphics[width=0.96\linewidth]{#1}}

\subsection{Datasets and evaluation protocol}

We use \textbf{Imagenet}~2012~\cite{DSLLF09} %
and follow a recent setup~\cite{bharath2017low} previously introduced for low-shot learning. The 1000 Imagenet classes are split randomly into two groups, each containing base and novel classes.
Group 1 (193 base and 300 novel classes) is used for hyper-parameter tuning and group~2 (196+311 classes) for testing with fixed hyper-parameters. We assume the full Imagenet training data is available for the base classes. For the novel classes, only $n$ images per class are available for training. Similar to~\cite{bharath2017low} the subset of $n$ images is drawn randomly and the random selection is performed 5 times with different random seeds. 

As a large source of unlabelled images, we use the \textbf{YFCC100M} dataset~\cite{TFENPBL16}. It consists of 99~million representative images from the Flickr photo sharing site\footnote{Of the 100M original files, some are videos and some are not available anymore. We replace them with uniform white images.}.  Note that some works have used this dataset with tags or GPS metadata as weak supervision~\cite{JMJV16}.

\paragraph{Learning the image descriptors. }
We use the 50-layer Resnet trained by Hariharan \etal~\cite{bharath2017low} on all base classes (group 1 + group 2), to ensure that the description calculation has never seen any image of the novel classes. We run the CNN on all images, and 
extract a 2048-dim vector from the 49th layer, just before the last fully connected layer. This descriptor is used directly as input for the logistic regression. For the diffusion, we PCA-reduce the feature vector to 256 dimensions and L2-normalize it, which is standard in prior works on unsupervised image matching with pre-learned image representations~\cite{BSCL14,TSJ16}.

\paragraph{Performance measure and baseline}

In a given group (1 or 2), we classify the Imagenet validation images from \emph{both} the base and novel classes, and measure the top-5 accuracy. Therefore the class distribution is heavily unbalanced. Since the seed images are drawn randomly, we repeat the random draws 5 times with different random seeds and average the obtained top-5 accuracy (the {\std{xx}} notation gives the standard deviation).

The baseline is a logistic regression applied on the labelled points. We employ a per-class image sampling strategy to circumvent the unbalanced number of examples per class. We optimize the learning rate, batch size and L2 regularization factor of the logistic regression on the group~1 images. It is worth noticing that our baseline outperforms the reported state of the art in this setting.

\paragraph{Background images for diffusion}

We consider the following sets of background images: 
\begin{enumerate}
  \setlength{\parskip}{0pt}
  \setlength{\parsep}{0pt}
\item{\em None:} the diffusion is directly from the seed images to the test images;
\item\emph{In-domain setting:}
	the background images are the Imagenet training image from the novel classes, but without labels. This corresponds to a use case where all images are known to belong to a set of classes, but only a subset of them have been labelled; 
\item\emph{Out-of-domain setting:} 
        the $\ndis$ background images are taken from YFCC100M. We denote this setting by F100k, F1M, F10M or F100M, depending on the number of images we use (\eg, we note  F1M for $\ndis=10^6$). This corresponds to a more challenging setting where we have no prior knowledge about the image used in the diffusion. 
\end{enumerate}

\subsection{Parameters of diffusion} 

\begin{table}[t]
\begin{center}
{\small
\begin{tabular}{|l|r@{\hspace{7pt}}r@{\hspace{7pt}}r@{\hspace{7pt}}|}
\hline
background & none & F1M & Imagenet  \\
\hline
\multicolumn{4}{|c|}{\textit{edge weighting}} \\
\hline
{\bf constant}     & 62.7\std{0.68} & 65.4\std{0.55} &  73.3\std{0.72} \\ 
Gaussian weighting*& 62.7\std{0.66} & 65.4\std{0.58} &  73.6\std{0.71} \\
meaningful neighbors* & 62.7\std{0.68} & 40.0\std{0.20} &  73.6\std{0.62}\\
\hline
\multicolumn{4}{|c|}{\textit{$\eta$ operator}} \\
\hline
none               & 40.6\std{0.18} & 41.1\std{0.10} & 42.3\std{0.19} \\
Sinkhorn           & 61.1\std{0.69} & 56.8\std{0.50} & 72.3\std{0.72} \\
{\bf column-wise } & 62.7\std{0.68} & 65.4\std{0.55} & 73.3\std{0.72} \\
non-linear transform* $\Gamma_r$   & 62.7\std{0.68} & 65.4\std{0.55} & 73.3\std{0.72} \\
class frequency prior* & 62.7\std{0.66} & 65.4\std{0.60} & 73.3\std{0.65} \\
\hline
\end{tabular}}
\end{center}
\caption{\label{tab:edgenorm}
Variations on weighting for edges and normalization steps on iterates of $\L$. The tests are performed for $n=2$ and $k=30$, with 5 runs on the group 1 validation images. 
Variants that require a parameter (\eg, the $\sigma$ of the Gaussian weighting) are indicated with a ``*''. In this case we report only the best result, see the supplementary material for full results. 
In the rest of the paper, we use the variants indicated \textbf{in bold}, since they are simple and do not add any parameter. 
}
\end{table}

We compare a few settings of the diffusion algorithm as discussed in section \ref{sec:variations}. In all cases, we set the number of nearest neighbors to $k=30$ and evaluate with $n=2$. The nearest neighbors are computed with Faiss  ~\cite{johnson2017billion}, using the IVFFlat index. It computes exact distances but occasionally misses a few neighbors. %
\paragraph{Graph edge weighting.}
We experimented with different weightings for $\W_0$, that were proposed in the literature. We compared a constant weight, a Gaussian weighting~\cite{LC10,chapterLabelProp06}, (with $\sigma$ a hyper-parameter), and a weighting based on the ``meaningful neighbors'' proposal~\cite{ODL07}.

Table~\ref{tab:edgenorm} shows that results are remarkably independent of the weighting choice, which is why we set it to 1\footnote{Note that our parametric experiments use the set of baseline image descriptors used in the arXiv version of ~\cite{bharath2017low}, and the table compares all methods using those underlying features, so the results are not directly comparable with the rest of the paper.
}. 
The best normalization that can be applied to the $\L$ matrix is a simple column-wise L1 normalization. 
Thanks to the linear iteration formula, it can be applied at the end of the iterations.

\subsection{Large-scale diffusion}

\begin{figure}[t]
\centering
\includegraphics[width=0.85\linewidth]{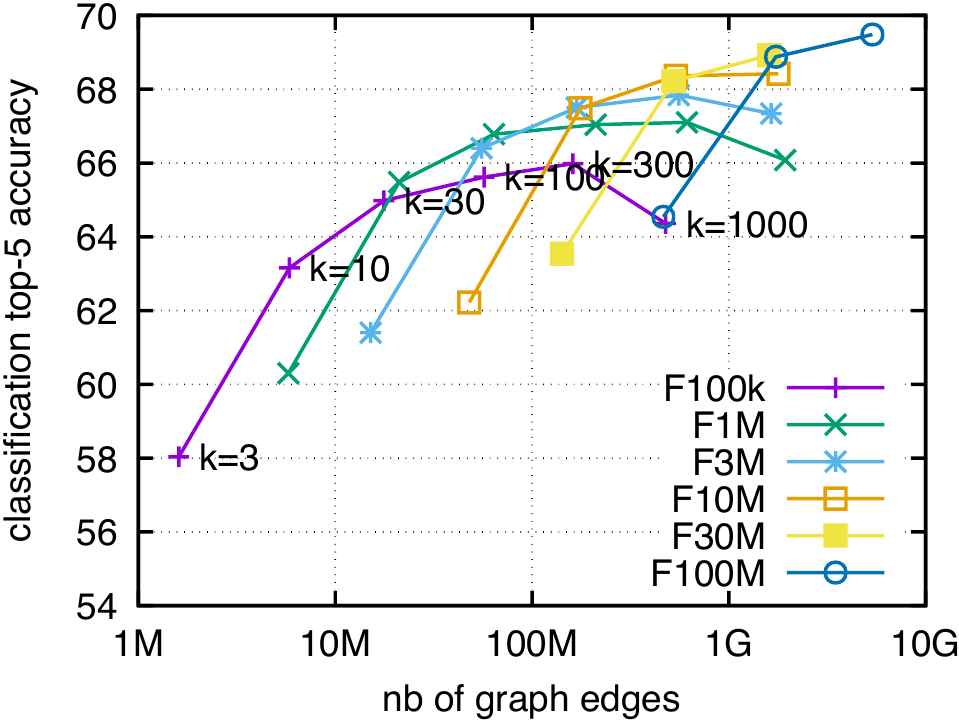}
\caption{\label{fig:largescale}
	Classification performance with $n=2$, with various settings of $k$ and $\ndis$, ordered by total number of edges (average of 5~test runs, with cross-validated number of iterations).
}
\end{figure}

Figure~\ref{fig:largescale} reports experiments by varying the number of background images $\ndis$ and the number $k$ of neighbors, for $n=2$. 
All the curves have an optimal point in terms of accuracy \emph{vs} computational cost at $k$=30. This may be a intrinsic property of the descriptor manifold. An additional number: before starting the diffusion iterations, with $k$=1000 and no background images (the best setting) we obtain an accuracy of 60.5\%. This is a knn-classifier and this is the fastest setting because the knn-graph does not need to be constructed nor stored. %

\subsection{Comparison with low-shot classifiers}
\begin{table}[t]
\begin{center}
{\small
\begin{tabular}{r|cc|cc|}
      & Hariharan                   & logistic       & in-domain      & diffusion      \\
 $n$  & \etal~\cite{bharath2017low} & regression     & diffusion      & + logistic     \\
\hline                                                                  
$1$   & 63.6                        & 60.4\std{0.78} & 69.7\std{0.86} & 69.76\std{0.88} \\ 
$2$   & 71.5                        & 68.8\std{0.82} & 75.4\std{0.64} & 75.60\std{0.69} \\
$5$   & 80.0                        & 79.1\std{0.35} & 79.9\std{0.17} & 81.35\std{0.22} \\
$10$  & 83.3                        & 83.4\std{0.16} & 82.1\std{0.14} & 84.56\std{0.12} \\
$20$  & 85.2                        & 86.0\std{0.15} & 83.6\std{0.12} & 86.72\std{0.09} \\
\hline\end{tabular}}
\end{center}
\caption{\label{tab:comparelogreg_indomain}
	\textbf{In-domain diffusion on Imagenet}: We compare against logistic regression and a recent low-shot learning technique~\cite{bharath2017low} on this benchmark. Results are reported with $k=30$ for diffusion.
}
\end{table}

\begin{table*}[t]
~\hfill
{\small
\begin{tabular}{r|cccc|c|cc|c}
      & \multicolumn{4}{c|}{out-of-domain diffusion}                        & logistic       & \multicolumn{2}{c|}{diffusion+logistic} & Hariharan \\
 $n$  & none           & F1M            & F10M            & F100M           & regression     & +F10M          & + F100M        & \etal~\cite{bharath2017low}\\
\hline
$1$    &  58.5\std{0.52}   &  61.4\std{0.61}   & 62.7\std{0.76}  & 63.6\std{0.61}  & 60.4\std{0.78} & 63.3\std{0.73} & \textbf{64.0}\std{0.70} &  63.6 \\ 
$2$    &  63.6\std{0.60}   &  66.8\std{0.71}   & 68.4\std{0.74}  & 69.5\std{0.60}  & 68.8\std{0.82} & 70.6\std{0.80} & 71.1\std{0.82} &  \textbf{71.5} \\ 
$5$    &  69.0\std{0.46}   &  72.5\std{0.27}   & 74.0\std{0.35}  & 75.2\std{0.40}  & 79.1\std{0.35} & 79.4\std{0.34} & 79.7\std{0.38} &  \textbf{80.0} \\ 
$10$   &  73.9\std{0.15}   &  76.2\std{0.19}   & 77.4\std{0.31}  & 78.5\std{0.34}  & 83.4\std{0.16} & 83.6\std{0.13} & \textbf{83.9}\std{0.10} &  83.3 \\ 
$20$   &  78.0\std{0.15}   &  79.1\std{0.23}   & 80.0\std{0.27}  & 80.8\std{0.18}  & 86.0\std{0.15} & 86.2\std{0.12} & \textbf{86.3}\std{0.17} &  85.2 \\ 
\hline
\end{tabular} \smallskip} 
\hfill ~
\caption{\label{tab:comparelogreg}
	\textbf{Out-of-domain diffusion:} Comparison of classifiers for different values of $n$, with $k=30$ for the diffusion results. The ``none'' column indicates that the diffusion solely relies on the labelled images. The results of the rightmost column~\cite{bharath2017low} are state-of-the-art on this benchmark to our knowledge, generally outperforming the results of matching networks and model regression \cite{VBLW16,WH16} in this setting.
}
\end{table*}

We compare the performance of diffusion against the logistic baseline classifiers and a recent method of the state of the art~\cite{bharath2017low}, using the same features. 

\paragraph{In-domain scenario.} 
For low-shot learning ($n\le 5$), the in-domain diffusion outperforms the other methods by a large margin, see Table~\ref{tab:comparelogreg_indomain}. The combination with logistic regression is not very effective.

\paragraph{Out-of-domain diffusion.} 
Table~\ref{tab:comparelogreg} shows that the performance of diffusion is competitive only when 1 or 2 images are available per class. 
As stated in Section~\ref{sec:propagationmatrix}, we do not include the test points in the diffusion, which is standard for a classification setting. However, if we allow this, as in a fully transductive setting, we obtain a top-5 accuracy of~{\textbf{69.6}\%\std{0.68}} with $n=2$ with diffusion over F1M, \ie, on par with diffusion over F100M.

\paragraph{Classifier combination.}

We experimented with a very simple late fusion: to combine the scores of the two classifiers, we simply take a weighted average of their predictions (log-probabilities), and cross validate the weight factor. Both in the in-domain (Table~\ref{tab:comparelogreg_indomain}) and out-of-domain (Table~\ref{tab:comparelogreg}) cases, the results are significantly above the best of the two input classifiers. This shows that the logistic regression classifier and the diffusion classifier access different aspects of image collection.  We also experimented with more complicated combination methods, like using the graph edges as a regularizer during the logistic regression, which did not improve this result.

\paragraph{Comparison with the state of the art.} 
With the in-domain diffusion, we notice that our method outperforms the state-of-the-art result of~\cite{bharath2017low} and which, itself, outperforms or is closely competitive with \cite{VBLW16,WH16} in this setting. In the out-of-domain setting, out results are better only for $n$=1. However, their method is a complementary combination of a specific loss and a learned data augmentation procedure that is specifically tailored to the experimental setup with base and novel classes. In contrast, our diffusion procedure is generic and has only two parameters ($\ndis$ and $k$).  Note that the out-of-domain setting is comparable with the standard low-shot setting, because the unlabeled images from F100M  are generic, and have nothing to do with Imagenet; and because the neighbor construction and diffusion are efficient enough to be run on a single workstation.

\subsection{Complexity: Runtime and memory}

\begin{table}[t!]
\begin{center}
{\small
\begin{tabular}{l|r@{\hspace{5pt}}r@{\hspace{5pt}}r@{\hspace{5pt}}r@{\hspace{5pt}}}
background        & none & F1M & F10M & F100M \\
\cline{1-5}
optimal iteration &    2   & 3     & 4    & 5    \\
timing: graph completion &  2m57s & 8m36s & 40m41s & 4h08m \\
timing: diffusion         &  4.4s  & 19s   & 3m44s  & 54m \\
\hline
\end{tabular}}
\end{center}
\caption{Timings for the different steps on a 24-core 2.5GHz machine, for a varying number of unlabelled images from YFCC100M. Note, the timing of 4h08m for graph completion over F100M takes only 23m when executed on 8 GPUs. %
\label{tab:timings}}
\end{table}

We measured the run-times of the different steps involved in diffusion process and report them in Table~\ref{tab:timings}.
The graph construction time is linear in~$\ndis$, thanks to the pre-computation of the graph matrix for the background images (see Section~\ref{sec:complexity}). 
For comparison, training the logistic regression takes between 2m27s and 12m, depending on the cross-validated parameters. %

In terms of memory usage, the biggest F100M experiments need to simultaneously keep in RAM a $\W$ matrix of 5.3~billion non-zero values (39.5~GiB), and  $\L_t$ and $\L_{t+1}$ (35.8~GiB, using slices of $C'=96$~columns). This is the main drawback of using diffusion. However Table~\ref{tab:comparelogreg} shows that restricting the diffusion to 10 million images already provides most of the gain, while dividing by an order of magnitude memory and computational complexity.

\subsection{Analysis of the diffusion process}
\label{sec:nonzeros}

\begin{figure}[t]
\includegraphics[width=0.99\linewidth]{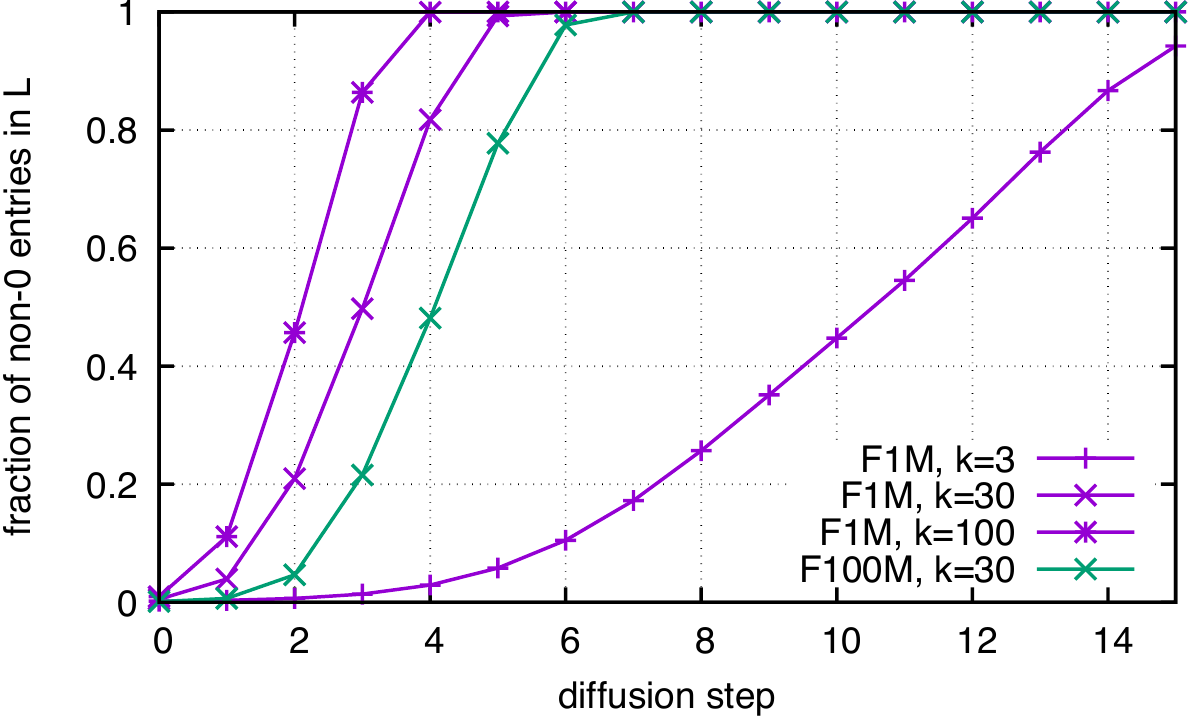} \medskip \\
\includegraphics[width=0.99\linewidth]{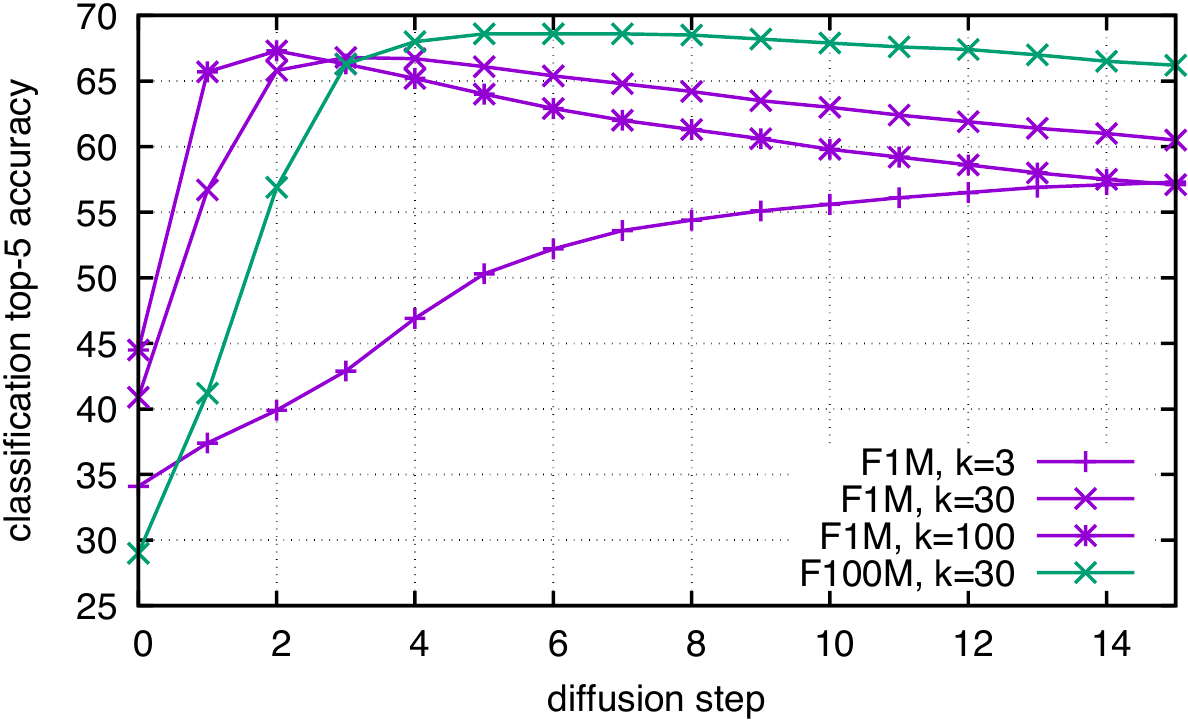} 
\caption{\label{fig:nonzero}
Statistics over iterations, for $n=2$. \emph{Top}: Rate of non-zero element in the matrix $\L$. \emph{Bottom}: corresponding accuracy.
}
\vspace{-8pt}
\end{figure}

We discuss how fast $\L$ ``fills up'' (it is dense after a few iterations). 
We consider the rate of nodes reached by the diffusion process: we consider very large graphs, few seeds and a relatively small graph degree. 
Figure~\ref{fig:nonzero} measures the sparsity of the matrix $\L$ (on one run of validation), which indicates the rate of (label, image) tuples that have not been attained by the diffusion process at each diffusion step.
While the graph is not necessarily fully connected, we observe that most images can be reached by all labels in practice. 

The fraction of nodes reached by all labeled points grows rapidly and converges to a value close to 1 in a few iterations when $k \geq 10$. 
In order to relate this observation with the performance attained along iterations, it is interesting to compare what happens in this plot to the one on the right. 
The plot on the right shows that the iteration number at which the matrix close to 1 is similar to the iteration at which accuracy is maximal, as selected by cross-validation. The maximum occurs later if $\ndis$ is larger and when $k$ is smaller. Note also that early stopping is important.

\newcommand{\igimnet}[4]{%
\raisebox{1mm}
{\framebox{
	\begin{minipage}[b]{1.9cm}%
	~\medskip \newline \small{#4} \newline \medskip ~%
	\end{minipage}}}%
}

\newcommand{\igtest}[4]{%
\raisebox{1mm}
{\framebox{
	\begin{minipage}[b]{1.9cm}%
	~\medskip \newline \footnotesize{\em (#4)} \newline \medskip ~%
	\end{minipage}}}%
}

\newcommand{\igflickr}[4]{%
\rotatebox{90}{\tiny{#4}}%
\includegraphics[width=1.9cm]{figs/images/#1/#3}%
}

\newcommand{\igimnetX}[4]{%
\begin{tabular}[b]{@{}c@{}}
\includegraphics[width=1.9cm]{figs/images/#1/#3}\\
\small{#4} \\
\end{tabular}%
}

\newcommand{\igtestX}[4]{%
\begin{tabular}[b]{@{}c@{}}
\includegraphics[width=1.9cm]{figs/images/#1/#3}\\
\small{\emph{(#4)}} \\
\end{tabular}%
}

\newcommand{\igflickrX}[4]{%
\begin{tabular}[b]{@{}c@{}}
\rotatebox{90}{\tiny{#4}}%
\includegraphics[width=1.9cm]{figs/images/#1/#3}\\
-- \\
\end{tabular}%
}

\begin{figure*}[t]
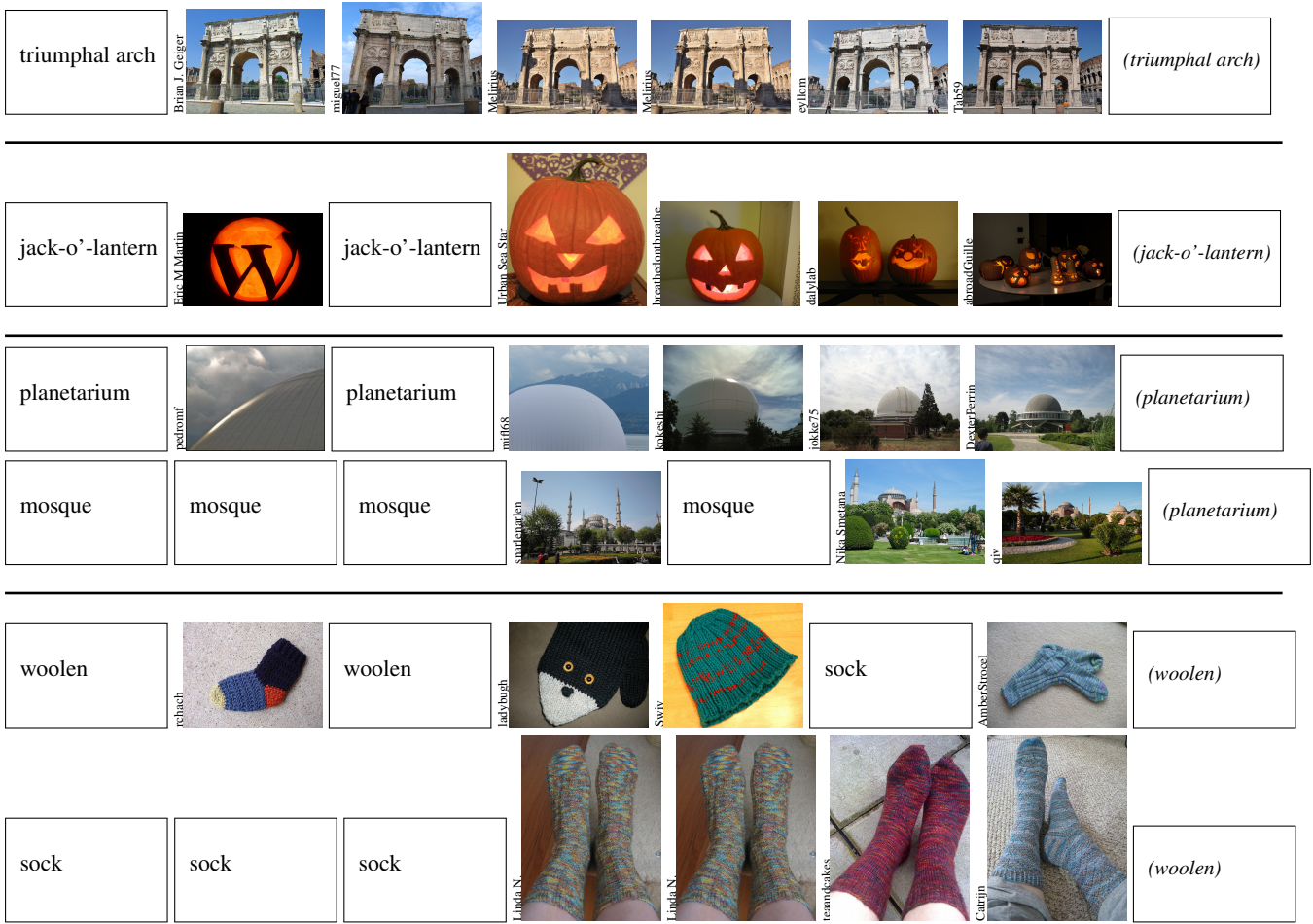

\hspace*{-2ex}
\begin{tabular}{l}
\input{figs/images/478/gt_list_user.tex} \\
\rule{\linewidth}{1pt} \\
\input{figs/images/23375/gt_list_user.tex} \\
\rule{\linewidth}{1pt} \\
\input{figs/images/1762/gt_list_user.tex} \\
\input{figs/images/1762/found_list_user.tex} \\
\rule{\linewidth}{1pt} \\
\input{figs/images/19790/gt_list_user.tex} \\
\input{figs/images/19790/found_list_user.tex} \\
\end{tabular}
\caption{\label{fig:images}
    Images visited during the diffusion process from a seed (\emph left) to the test image (\emph{right}). We give  ground-truth class for Imagenet images (test images marked by parentheses). The first two rows are classified correctly. The two bottom ones are failure cases. 
    Imagenet images are not shown for copyright reasons, but the labels are shown. 
    For YFCC100M images, we provide the Flickr id of their creators.
}
\end{figure*}

\subsection{Qualitative results}

Figure~\ref{fig:images} shows paths between a seed image and test images, which gives a partial view of the diffusion. Given a class, we backtrack the path: for a given node (image) and iteration $i$, we look up the preceding node that contributed most to the weight in $\L_i$ that node at that iteration. At iteration 0, the backtracking process always ends in a source node. %
Each row of the figure is one such paths. For a test image (right), we show the path for the ground-truth class and that for the found class, or a single row for both when the image is classified correctly. Note that the preceding node can be the image itself, since the diagonal of the $\W$ matrix is not set to 0. %
Thanks to the size of the dataset, the paths are ``smooth'': they evolve through similar images.

\section{Conclusion}
\label{sec:conclusion}
\vspace*{-1ex}
We experimented on large-scale label propagation for low-shot learning.  Unsurprisingly, we have found that 
performing diffusion over images from the same domain works much better than images from a different domain. 
We clearly observe that, as the number of images over which we diffuse grows, the accuracy steadily improve. %
The main performance factor is the total number of edges, which also reasonably reflects the complexity.  
We also report neutral results for most sophisticated variants, for instance %
we show that edge weights are not useful.  Furthermore, labeled images should be included in the diffusion process and not just used as sources, \ie, not enforced to keep their label. 

The main outcome of our study is to show that diffusion over a large image set is superior to state-of-the-art methods for low-shot learning when very few labels are available. Interestingly, late-fusion with a standard classifier's result is effective. This shows the complementary of the approaches, and suggests that it could be combined with forthcoming methods for low-short learning.  

When more labels are available, simple logistic regression becomes superior to the methods we describe (and to other state of the art low-shot learning methods).  However, we note that there are many circumstances where even a few labels per class are more difficult to get than building (and then keeping) a graph over unlabeled data.   For example, if there are a large number of ``tail'' classes which we will need to classify, a few examples per class can multiply to many labels.   
In these cases diffusion combined with logistic regression is the best method.
The code to reproduce our results is available at \scalebox{0.67}{\texttt{https://github.com/facebookresearch/low-shot-with-diffusion}}.

{\small
\bibliographystyle{abbrv}
\bibliography{egbib}
}

\section*{Appendix}

\appendix

\iffalse

\documentclass[10pt,twocolumn]{article}

\usepackage{cvpr}

\usepackage{times}
\usepackage{epsfig}
\usepackage{graphicx}
\usepackage{amsmath}
\usepackage{amssymb}
\usepackage{rotating}
\usepackage{color}
\usepackage{framed}

\def \ie {\emph{i.e.}}
\def \eg {\emph{e.g.}}
\def \etal {\emph{et al.}}
\def\cvprPaperID{1590} %
\cvprfinalcopy

\ifcvprfinal
        \pagestyle{empty}
        \def\httilde{\mbox{\tt\raisebox{?}{?}x{-.5ex}{\symbol{126}}}}
\fi

\def \W {\mathbf{W}}
\renewcommand{\L}{\mathbf{L}} 

\newcommand{\ndis}{n_\mathrm{B}}
\newcommand{\nlab}{n_\mathrm{L}}
\newcommand{\std}[1]{\footnotesize{$\pm$#1}}

\definecolor{darkgreen}{RGB}{0, 140, 0}
\newcommand{\rv}[1]{{\color{darkgreen}[\textbf{Rv}:#1]}}
\newcommand{\arthur}[1]{{\color{red}[\textbf{Arthur}:#1]}}
\newcommand{\matthijs}[1]{{\color{blue}[\textbf{Matthijs}:#1]}}
\usepackage[pagebackref=true,breaklinks=true,letterpaper=true,colorlinks,bookmarks=false]{hyperref}

{\renewcommand{\baselinestretch}{1.05}

\begin{document}

\title{Supplementary material for: \\Low-shot learning with large-scale diffusion}

\author{Matthijs Douze\textsuperscript{$\dagger$}, 
Arthur Szlam\textsuperscript{$\dagger$}, 
Bharath Hariharan\textsuperscript{$\dagger$}\thanks{This work was carried out while B. Hariharan was post-doc at FAIR.}~, 
Herv\'e J\'egou\textsuperscript{$\dagger$}\\
\textsuperscript{$\dagger$}Facebook AI Research\\
\textsuperscript{*}Cornell University\\
}

\maketitle

\else

\appendix

\fi

We present several additional results and details to complement the paper. %
Section~\ref{sec:evalnovel} reports another evaluation protocol, which restricts the evaluation to novel classes. Sections~\ref{sec:params} and~\ref{sec:lfweights} are parametric evaluations. Section~\ref{sec:Wblocks} gives some details about the graph computation.

\section{Evaluation results on novel classes}
\label{sec:evalnovel}

In the main paper, we evaluated the search performance on all the test images from group 2. The performance restricted to only the novel classes is also reported in prior work~\cite{bharath2017low} using a combination of classifiers. 
Table~\ref{tab:comparelogregnovel} shows the results in this setting. 

\begin{table*}[h]
\centering
{
\hspace*{-5mm}
\begin{tabular}{r|cccc|c|c|cc|c}
\multicolumn{1}{c}{}	   & \multicolumn{4}{c|}{out-of-domain diffusion}  & in-domain    & logistic &  \multicolumn{2}{c|}{combined} & best reported\\
 $n$      & none & F1M & F10M & F100M & (Imagenet) &  regression  & +F10M & + F100M & results \cite{bharath2017low} \\
\hline
\hline
$1$    & 39.4\std{0.85} &  43.9\std{0.96} &  46.3\std{1.28} & 47.6\std{1.09} & 57.7\std{1.28} & 42.6\std{1.31} & 46.5\std{1.23} & \textbf{47.9}\std{1.18} & 45.1 \\ 
$2$    & 47.8\std{0.94} &  52.7\std{1.14} &  55.2\std{1.21} & 57.0\std{1.05} & 66.9\std{1.06} & 54.4\std{1.29} & 57.5\std{1.34} & 58.4\std{1.29} & \textbf{58.8} \\ 
$5$    & 56.8\std{0.73} &  62.2\std{0.44} &  64.6\std{0.57} & 66.3\std{0.68} & 73.8\std{0.29} & 71.4\std{0.54} & 71.9\std{0.55} & 72.3\std{0.58} & \textbf{72.7} \\  
$10$   & 64.9\std{0.28} &  68.0\std{0.33} &  69.9\std{0.47} & 71.7\std{0.54} & 77.6\std{0.23} & 78.6\std{0.27} & 78.7\std{0.21} & \textbf{79.2}\std{0.14} & 79.1 \\  
$20$   & 71.4\std{0.26} &  72.7\std{0.40} &  74.1\std{0.38} & 75.3\std{0.29} & 80.0\std{0.21} & 82.9\std{0.20} & 83.0\std{0.15} & \textbf{83.2}\std{0.22} & 82.6 \\  
\hline
\end{tabular}}
\smallskip
\caption{\label{tab:comparelogregnovel}
	Comparison of classifiers for different values of $n$, with $k=30$ for the diffusion results, evaluating \emph{only on novel classes} using the two different sets of features that we consider.
}
\medskip
\end{table*}

As to be expected, the results reported in these tables are inferior to those obtained in the setup where all test images are classified. This is because the novel classes are harder to classify than the base classes. Otherwise the ordering of the methods is preserved and the conclusions identical. The diffusion is effective in the low-shot regime and is, by itself, better than the state of the art by a large margin when only one example is available. The combination with late fusion significantly outperforms the state of the art, even in the out-of-domain setup.

\section{Details of the parametric evaluation}
\label{sec:params}

\begin{figure*}
\begin{minipage}{0.48\linewidth}
\centering
Gaussian weighting  \\
\includegraphics[width=\linewidth]{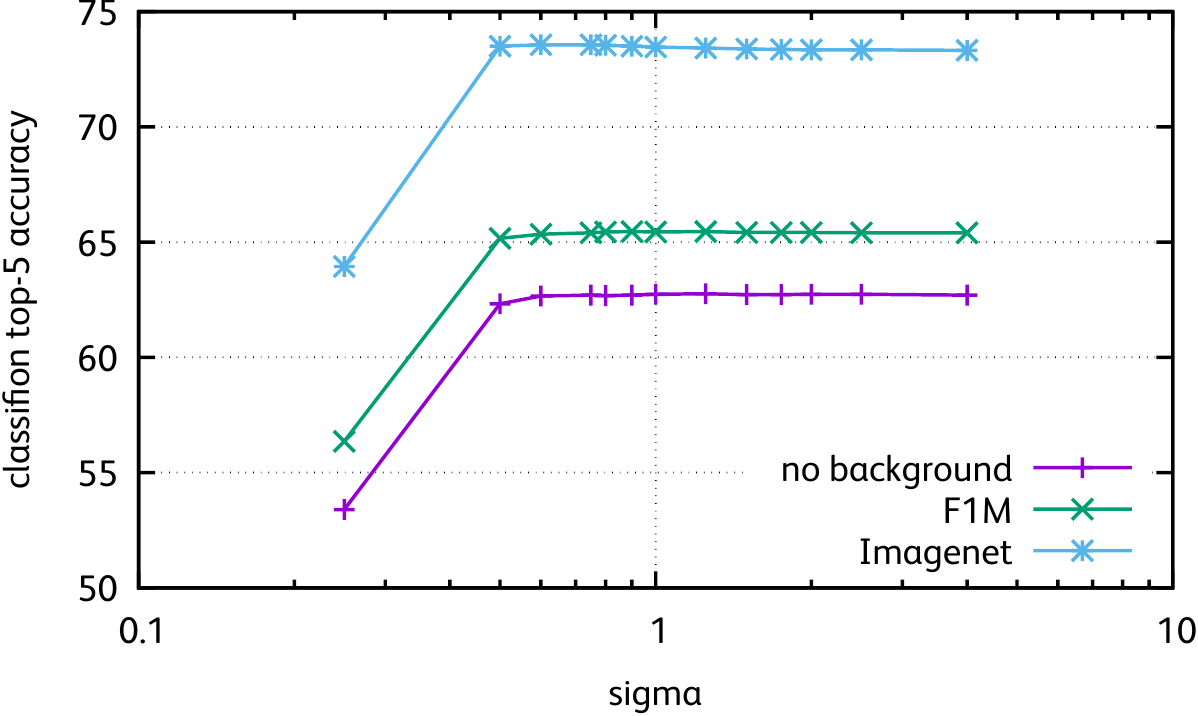}
\end{minipage}
\hfill
\begin{minipage}{0.48\linewidth}
\centering
Meaningful neighbors model \\
\includegraphics[width=\linewidth]{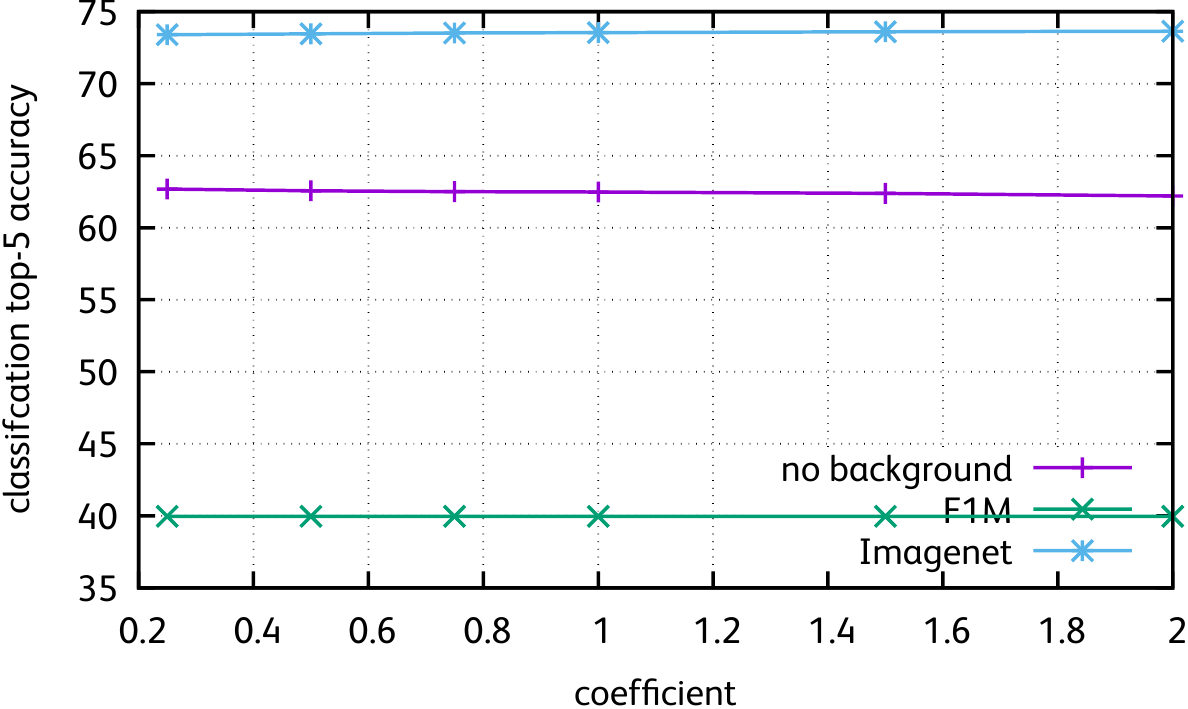} 
\end{minipage}
\bigskip 
\\
\begin{minipage}{0.48\linewidth}
\centering
$\Gamma_r$ normalization \\
\includegraphics[width=\linewidth]{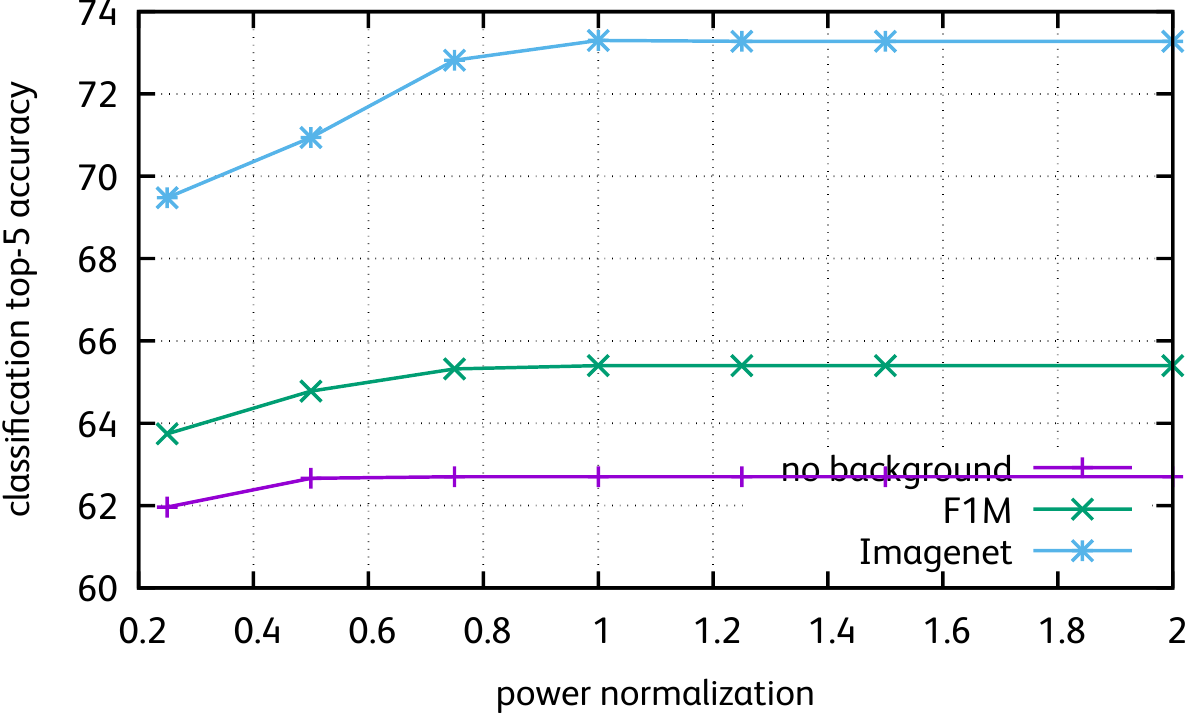} 
\end{minipage}
\hfill
\begin{minipage}{0.48\linewidth}
\centering
Normalization with class weights \\
\includegraphics[width=\linewidth]{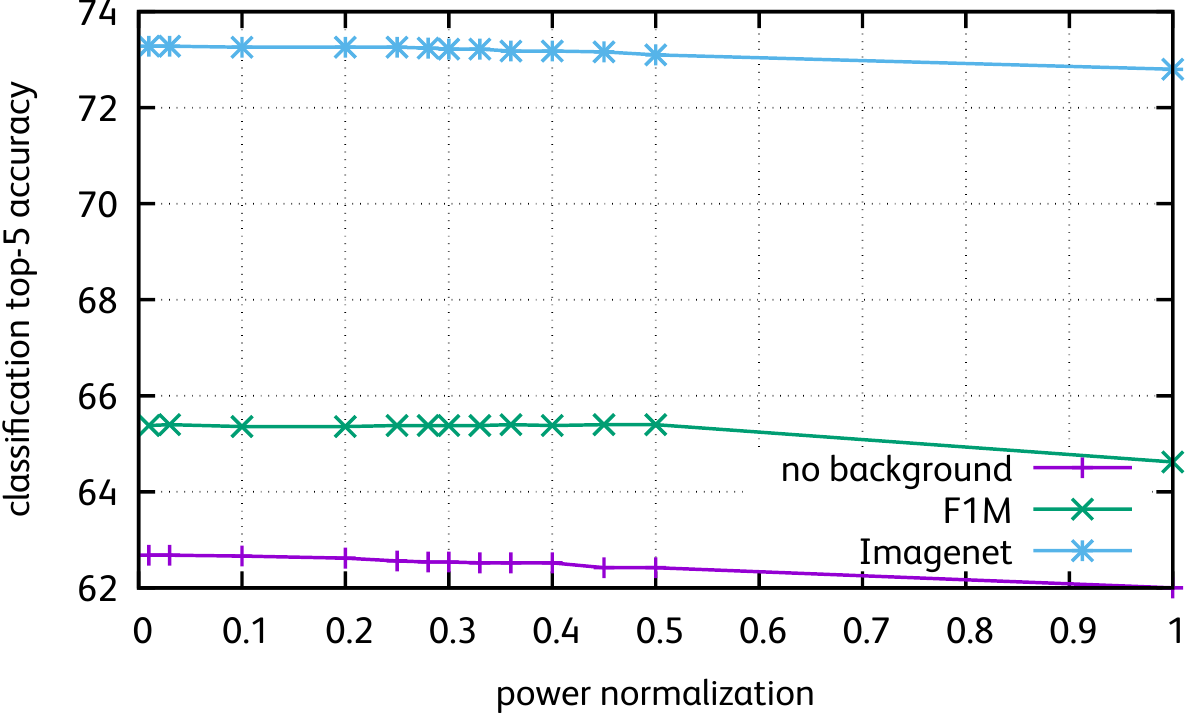} 
\end{minipage}
\bigskip
\caption{\label{fig:normalizations}
	Evaluation of edge weighting (top) and matrix normalizations (bottom) used in the diffusion. The common settings are: $k=30$, $n=2$, evaluation is averaged over 5 runs on the validation set (group 1), and we select the best iteration. 
}
\end{figure*}

In the paper we reported results for the edge weighting and graph normalization with the best parameter setting. Here, we report results for all parameters\footnote{Note that our parametric experiments use the set of baseline image descriptors used in the arXiv version of the paper by Barath \etal~\cite{bharath2017low}, and the figure compares all methods using those underlying features. Therefore the results are not directly comparable with the rest of the paper.
}.
We evaluate the following edge weightings (Figure~\ref{fig:normalizations}, first row): 
\begin{itemize}
\item
	Gaussian weighting. The edge weight is~$e^{-x^2/\sigma^2}$ with $x$ the distance between the edge nodes. Note that $\sigma\rightarrow\infty$ corresponds to a constant weighting;
\item
	Weighting based on the ``meaningful neighbors'' proposal~\cite{ODL07}. It relies on an exponential fit of neighbor distances. For a given graph node, for the neighbor $i$ of its list of results, the weight is $s(1-e^{-\lambda s})^k$, where $s$ is the distance, remapped linearly to $[0,1]$ so that the first neighbor has $s=1$ and the $k$\textsuperscript{th} neighbor has $s=0$. We vary parameter $\lambda$ in the plot. 
\end{itemize}

We also report results for different normalizations of the matrix $\mathbf{L}$. In Figure~\ref{fig:normalizations} (second row), we compare:
\begin{itemize}
\item
	The non-linear $\Gamma_r$ normalization, all elements of $\mathbf{L}$ are raised to a power $r$. We vary the parameter $r$, and $r=1$ corresponds to the identity transform;
\item	
	We classify all images in a graph with a logistic regression classifier. We use the predicted frequency of each class over the whole graph, and raise it to some power (the parameter) to reduce or increase its peakiness. This choice is inspired by the Markov Clustering Algorithm~\cite{EDO2002}. This gives a normalization factor that we enforce for each column of $\mathbf{L}$, instead of the default uniform distribution. 
\end{itemize}

The conclusion of these experiments is that these variants do not improve over constant weights and a standard diffusion, most of them having a neutral effect. Therefore, we conclude that the diffusion process mostly depends on the topology of the graph.

\section{Late fusion weights}
\label{sec:lfweights}

\begin{figure}
\begin{center}
\includegraphics[width=\linewidth]{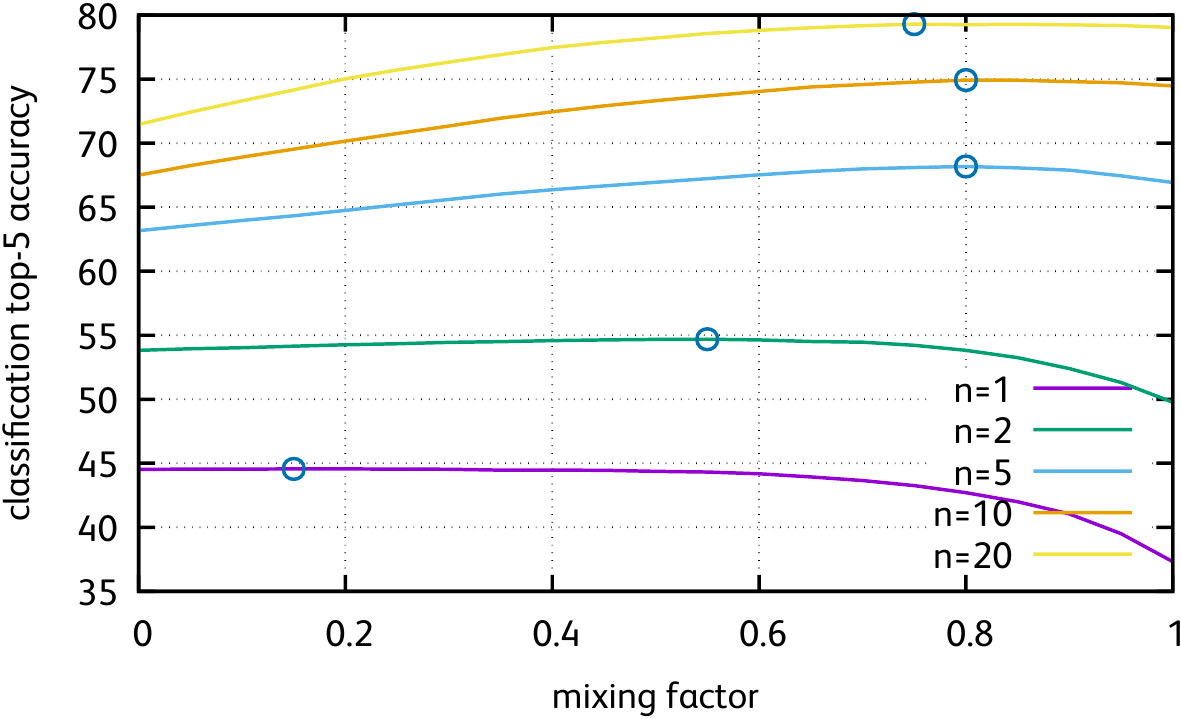}
\end{center}
\caption{\label{fig:mixweight}
	Performance as a function of the late fusion weight $a_n$ on the validation (group 1) images, averaged over 5 runs. 
	A weight of 0 is pure diffusion, 1 is pure logistic regression. The mixing factors that are selected for test are indicacted with circles.
}
\end{figure}

Let denote by $l_i^\mathrm{logreg}$ and $l_i^\mathrm{dif} \in [0,1]^C$ the distributions over classes returned by the two classifiers for image $i$. We fuse the loglikehood by a weighted average, which amounts to retrieving the top-5 class prediction as those maximizing 
\begin{equation}
a_n\log(l_{ic}^\mathrm{logreg}) + (1-a_n)\log(l_{ic}^\mathrm{dif}), 
\end{equation}
where $a_n$ is the optimal mixing coefficient for $n$ seed points, as found by cross-validation. 

Figure~\ref{fig:mixweight} shows these optimal mixing factors. Since the logistic regression is better at classifying with many training examples, the parameter $a_n$ increases with $n$.

\section{Computation of the $\W_0$ blocks} 
\label{sec:Wblocks}

As stated in the paper,  we need to compute the 4 blocks of the matrix $\W_0$:
\begin{equation}
\W_0 = \left[
\begin{array}{cc}
\W_\mathrm{LL} & \W_\mathrm{LB} \\
\W_\mathrm{BL} & \W_\mathrm{BB} \\
\end{array}
\right] \in \{0,1\}^{(\nlab+\ndis)\times (\nlab+\ndis)},
\end{equation}
where, usually, $\nlab \ll \ndis$. Each block requires to perform a $k$-nearest neighbor search. We employ the Faiss library\footnote{\url{http://github.com/facebookresearch/faiss}} optimized for this task~\cite{johnson2017billion}, and use it as follows:
\begin{itemize}
\item
	$\W_\mathrm{BB}\in \{0,1\}^{\ndis\times \ndis}$: we use a Faiss index referred to as ``IVFFlat''. The accuracy-speed compromise is controlled by a parameter giving the number of inverted lists visited at search time. We adopted a relatively high probe setting (256) to guarantee that most of the actual neighbors are retrieved. With the recommended settings of Faiss, the complexity of one search is proportional to $d\sqrt{\ndis}$, so the total complexity is $\mathcal{O}(d\ndis^{1.5})$. This is super-linear with respect to $\ndis$, but it is still relatively efficient (see our timings) and performed off-line;
\item 
	$\W_\mathrm{LB}\in \{0,1\}^{\nlab\times \ndis}$: we re-use the same index to do $\nlab$ similarity search operations, this time using only $\mathcal{O}(d\nlab\sqrt{\ndis})$;
\item 
	$\W_\mathrm{BL}\in \{0,1\}^{\ndis\times \nlab}$: we need to index on the seed image descriptors. We found that in practice, constructing an index on these images is at best 1.4$\times$ faster than brute-force search. Therefore, we use brute-force search in this case, which if of order $\mathcal{O}(d\ndis \nlab)$;
\item
	$\W_\mathrm{BB}\in \{0,1\}^{\nlab\times \nlab}$: it has a negligible complexity.
\end{itemize}	

The fusion of the result lists $[\W_\mathrm{LL}~~\W_\mathrm{LB}]$ and $[\W_\mathrm{BL}~~\W_\mathrm{BB}]$ to get $k$ results per row of $\W_0$ is done in a single pass and in a negligible amount of time. 
Therefore the dominant complexity is $\mathcal{O}(d\ndis \nlab)$. A typical breakdown of the timings for F100M is (in seconds):

\begin{center}
\begin{tabular}{l|rrrr}
Timings (s)       & $\W_\mathrm{BB}$ & $\W_\mathrm{LB}$ & $\W_\mathrm{BL}$ & $\W_\mathrm{LL}$\\
\hline
on CPU    & ---  & 65+2783 & 12003 & 32 \\
on 8 GPUs & 25929 & 732+64  & 533   & 1 \\
\end{tabular}
\end{center}

For $\W_\mathrm{LB}$ we decompose the timing into: loading of the precomputed IVFFlat index (and moving it to GPU if appropriate) and the actual computation of the neighbors.

\iffalse

{\small
\bibliographystyle{abbrv}
\bibliography{egbib}
}
\end{document}